\documentclass[conference]{IEEEtran}
\usepackage{times}
\pdfoutput=1
\IEEEoverridecommandlockouts
\usepackage{hyperref}

\usepackage{balance} %even columns on last page
\usepackage{xcolor,calc}
\usepackage[pdftex]{graphicx}
\usepackage{amsmath} 
\usepackage{amssymb}  
\usepackage[noadjust]{cite}
\usepackage{color}
\usepackage{enumerate}
\usepackage{subfig}
\usepackage{multirow}
\usepackage{mathtools}
\usepackage{booktabs}
\usepackage{epstopdf}
\usepackage{algcompatible}
\usepackage[linesnumbered,ruled,vlined]{algorithm2e}
\usepackage[normalem]{ulem}
\usepackage{url}
\usepackage{bm}
\usepackage{booktabs}

\SetKwInput{KwInit}{Initialization}                % Set the Input
\SetKwInput{KwInput}{Input}

\makeatletter
\newsavebox\myboxA
\newsavebox\myboxB
\newlength\mylenA

\newcommand*\xoverline[2][0.75]{%
    \sbox{\myboxA}{$\m@th#2$}%
    \setbox\myboxB\null% Phantom box
    \ht\myboxB=\ht\myboxA%
    \dp\myboxB=\dp\myboxA%
    \wd\myboxB=#1\wd\myboxA% Scale phantom
    \sbox\myboxB{$\m@th\overline{\copy\myboxB}$}%  Overlined phantom
    \setlength\mylenA{\the\wd\myboxA}%   calc width diff
    \addtolength\mylenA{-\the\wd\myboxB}%
    \ifdim\wd\myboxB<\wd\myboxA%
       \rlap{\hskip 0.5\mylenA\usebox\myboxB}{\usebox\myboxA}%
    \else
        \hskip -0.5\mylenA\rlap{\usebox\myboxA}{\hskip 0.5\mylenA\usebox\myboxB}%
    \fi}
\makeatother

\usepackage{tabularx}
\usepackage{booktabs}

\newtheorem{assumption}{Assumption}

\begin{document}

% paper title
\title{Safe Human-Robot Collaborative Transportation via \\ Trust-Driven Role Adaptation}

% You will get a Paper-ID when submitting a pdf file to the conference system
\author{Tony Zheng$^\star$, Monimoy Bujarbaruah$^\star$, Yvonne R. St{\"u}rz, Francesco Borrelli % <-this % stops a space
\thanks{$^\star$authors contributed equally; E-mails: \tt\scriptsize{\{tony\_zheng, monimoyb, y.stuerz, fborrelli\}@berkeley.edu}}
}

%\author{\authorblockN{Michael Shell}
%\authorblockA{School of Electrical and\\Computer Engineering\\
%Georgia Institute of Technology\\
%Atlanta, Georgia 30332--0250\\
%Email: mshell@ece.gatech.edu}
%\and
%\authorblockN{Homer Simpson}
%\authorblockA{Twentieth Century Fox\\
%Springfield, USA\\
%Email: homer@thesimpsons.com}
%\and
%\authorblockN{James Kirk\\ and Montgomery Scott}
%\authorblockA{Starfleet Academy\\
%San Francisco, California 96678-2391\\
%Telephone: (800) 555--1212\\
%Fax: (888) 555--1212}}

% avoiding spaces at the end of the author lines is not a problem with
% conference papers because we don't use \thanks or \IEEEmembership

% for over three affiliations, or if they all won't fit within the width
% of the page, use this alternative format:
% 
%\author{\authorblockN{Michael Shell\authorrefmark{1},
%Homer Simpson\authorrefmark{2},
%James Kirk\authorrefmark{3}, 
%Montgomery Scott\authorrefmark{3} and
%Eldon Tyrell\authorrefmark{4}}
%\authorblockA{\authorrefmark{1}School of Electrical and Computer Engineering\\
%Georgia Institute of Technology,
%Atlanta, Georgia 30332--0250\\ Email: mshell@ece.gatech.edu}
%\authorblockA{\authorrefmark{2}Twentieth Century Fox, Springfield, USA\\
%Email: homer@thesimpsons.com}
%\authorblockA{\authorrefmark{3}Starfleet Academy, San Francisco, California 96678-2391\\
%Telephone: (800) 555--1212, Fax: (888) 555--1212}
%\authorblockA{\authorrefmark{4}Tyrell Inc., 123 Replicant Street, Los Angeles, California 90210--4321}}

\maketitle

\begin{abstract}
We study  a human-robot collaborative transportation task in presence of obstacles. The task for each agent is to carry a rigid object to a common target position, while safely avoiding obstacles and satisfying the compliance and actuation constraints of the other agent. Human and robot do not share the local view of the environment. The human policy either assists the robot when they deem the robot actions safe based on their perception of the environment, or actively leads the task. 

% The robot only models the assisting behavior of the human. The robot controller is only designed to assist the human. In particular,
Using estimated human inputs, the robot plans a trajectory for the transported object by solving a constrained finite time optimal control problem. Sensors on the robot measure the inputs applied by the human. The robot then appropriately applies a weighted combination of the human's applied and its own planned inputs, where the weights are chosen based on the robot's \emph{trust value} on its estimates of the human's inputs. This allows for a dynamic leader-follower role adaptation of the robot throughout the task. Furthermore, under a low value of trust, if the robot approaches any obstacle potentially unknown to the human, it triggers a safe stopping policy, maintaining safety of the system and signaling a required change in the human's intent. With experimental results, we demonstrate the efficacy of the proposed approach.
\end{abstract}

\IEEEpeerreviewmaketitle

\section{Introduction}
Human robot collaborative tasks have been a focus of major research work in robotics \cite{bauer2008human,jarrasse2014slaves, sheridan2016human}. For such tasks, roles of the agents are important, especially so in collaborative transportation. This is due to the fact that the transported object poses a compliance constraint that must be satisfied. Only follower or helper role of the robot can be seen in \cite{khatib1999mobile, kosuge2004human, maeda2001human, yokoyama2003cooperative}. In these works, the human knows the full environment and is the lead planner in the task. The robot follows the human by minimizing its felt forces and torques, and has no planning algorithms of its own. However, such fixed role assignment can be debilitating in situations when both agents have partial environment information, or if the human wants to lower their efforts in the task. Therefore shared and/or switching roles are introduced in \cite{evrard2009homotopy, oguz2010haptic, mortl2012role, beton2017leader, sadrfaridpour2018trust, kwon2019influencing, van2020adaptive}. In such switching role assignments, it is essential for the robot to make predictions of the human's intent from the human's observed behavior and then adapt its policy accordingly during the task. Such human intent prediction related work are also available in the literature \cite{freedy2007measurement, mainprice2013human, fisac2018probabilistically,bajcsy2017learning,yu2020estimation}. Obstacle avoidance in such human-robot collaborative tasks was studied in \cite{flacco2012depth, wang2013vision}, etc. However, to the best of our knowledge, the presence of unknown obstacles in the environment, inferring these obstacle positions from haptic feedback data and then explicitly incorporating the obstacle avoidance constraints in the robot's planning problem have not been addressed.  

% Collision avoidance studied in \cite{flacco2012depth, wang2013vision}. 

\begin{figure}[h!]
	\centering 	\includegraphics[width=\columnwidth]{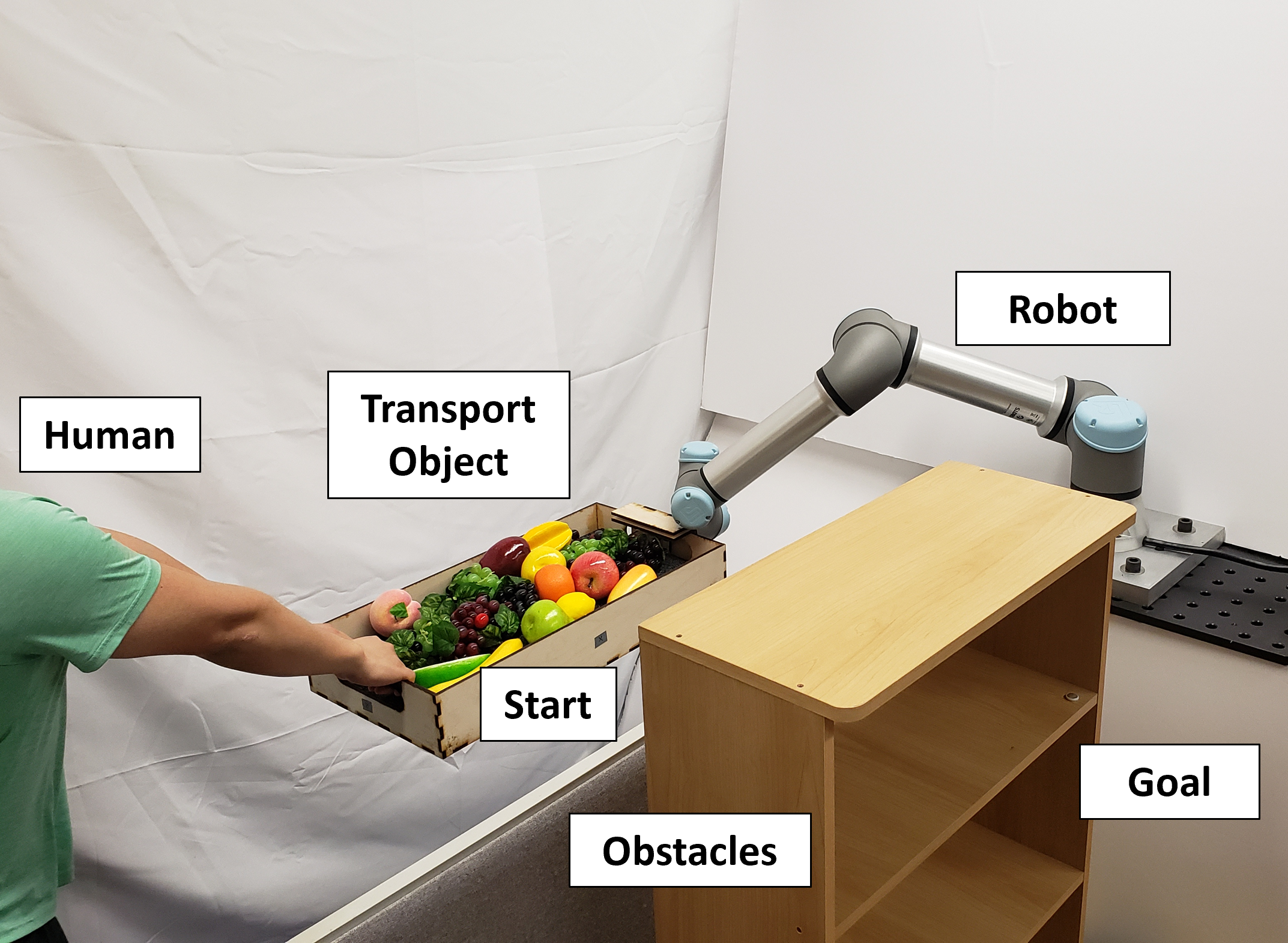}
	\caption{The considered experiment setup.}
	\label{fig:exp_space}    
\end{figure}

In this paper, we propose a Model Predictive Control (MPC) based strategy for a human-robot joint transportation task, as shown in Fig.~\ref{fig:exp_space}. The environment has obstacles partially known to each agent. The human's policy is allowed to be a combination of compliance and leadership, based on the human's intent during the task. The robot only estimates the compliant human behavior, and operates on a policy based on a computed \emph{trust value} and also its proximity to obstacles. This allows for a dynamic leader-follower role of the robot throughout the task, depending on the learned value of trust from applied human inputs. The trust is low if the actual human inputs differ highly from the robot's estimates, and vice versa. Our proposed framework can be summarized as:
\begin{itemize}
    \item We design a two mode policy for the robot. The first mode is the nominal operation mode, where the robot solves an MPC problem for its control synthesis. The cost function in the MPC optimization problem adapts based on the corrective inputs of the human to the robot's inputs. This enables the robot to plan trajectories that adapt with the human's behavior. 
    
    \item The control applied by the robot in the first mode is a function of the \emph{trust value}, similar to \cite{mortl2012role}. That is, after solving the MPC problem, the robot appropriately applies a weighted combination of the human's and its own planned actions, where the weights are adapted based on the deviation between robot's estimated and the actual human inputs.
    
    \item The second mode of the robot's policy is a safe stopping backup, which is triggered when the robot nears obstacles under a low value of trust on its estimated human's inputs. This safe stop mode enables the robot to decelerate the object, avoid collisions, and signal a required change in intent to the human via the haptic feedback. 
\end{itemize}
We highlight that the robot obtains a follower's role for low trust value, including safe stopping backup. On the other hand, it asserts a leader's role for high trust value, relying more on its MPC planned inputs. These leader-follower roles switch dynamically throughout the task as a function of the trust value. In Section~\ref{sec:numerics}, with experiments on a UR5e robot, we demonstrate the efficacy of our proposed approach. We present an experiment where with pre-assigned fixed roles the agents collide with obstacles, whereas a combination of trust-driven and safe stop policies manages to complete the task safely. 

\section{Problem Formulation}
In this section, we formulate the collaborative obstacle avoidance problem. We restrict ourselves to the case of two agents. The case of collaborative transportation with multiple agents is left as a subject of future research.
%%%%%%%%%%%%
\subsection{Environment Modeling}
Let the environment be contained within a set $\mathcal{X}$. In this work, we assume that the obstacles in the environment are static, although the proposed framework can be extended to dynamic obstacles. At any time step $t$, let the set of obstacle constraints known to the human and the robot (detected at $t$ and stored until $t$) be denoted by $\mathcal{C}_{h,t}$ and $\mathcal{C}_{r,t}$, respectively. We denote:
\begin{align*}
    &\mathcal{C}_{r,t} \cup \mathcal{C}_{h,t} = \mathcal{O}_t,~\forall t \leq T, 
\end{align*}
where $T \gg 0$ is the task duration limit and $\mathcal{O}_t$ is the set of obstacle constraints to be avoided at $t$ during the task. The approach proposed in this paper focuses on the challenging situation where no agent has the full information of all the detected obstacles in $\mathcal{O}_t$, i.e., $\mathcal{C}_{h,t} \subset \mathcal{O}_t$ and $\mathcal{C}_{r,t} \subset \mathcal{O}_t$. 

\subsection{System Modeling}
We model both the human and the robot transporting a three dimensional rigid object. Let $(\vec{I}_I, \vec{J}_I, \vec{K}_I)$ and $(\vec{I}_B, \vec{J}_B, \vec{K}_B)$ be the orthogonal unit bases vectors defining the inertial and the transported object fixed coordinate frames, respectively. Let $(X, Y, Z)$ be the position of the center of mass of the transported object in the inertial frame, $\vec{v}$ be the velocity of the center of mass relative to the inertial frame, expressed in the body-frame as
\begin{align}
    \vec{v} = v_x \vec{I}_B + v_y \vec{J}_B + v_z \vec{K}_B.
\end{align}
Furthermore, let the Euler angles $E = \begin{bmatrix} \psi & \theta & \phi \end{bmatrix}^\top$ be the roll, pitch, yaw angles describing the orientation of the body w.r.t. the inertial frame, and $\vec{\omega}_{B/I}$ be the angular velocity of the body-fixed frame w.r.t. the inertial frame, expressed in the body-fixed frame as
\begin{align}
    \vec{w}_{B/I} = w_x \vec{I}_B + \omega_y \vec{J}_B + \omega_z \vec{K}_B.
\end{align}
We denote $\dot{E} = W^{-1} \begin{bmatrix} \omega_x & \omega_y & \omega_z \end{bmatrix}^\top$, with matrix 
\begin{align*}
    W^{-1} = \frac{1}{\cos \theta} \begin{bmatrix} 0 & \sin \phi & \cos \phi \\ 0 & \cos \phi \cos \theta & -\sin \phi \cos \theta \\ \cos \theta & \sin \phi \sin \theta & \cos \phi \sin \theta \end{bmatrix}.
\end{align*}
Let $(F_x, F_y, F_z)$ be the force components along the inertial axes applied at the body's center of mass, $(\tau_x, \tau_y, \tau_z)$ are the torques about the body fixed axes, and $J$ be the moment of inertia of the body expressed in the body frame, given by $J = \mathrm{diag}( J_x, J_y, J_z)$. Then the equations of motion of the object transported are written as follows:
\begin{equation} \label{eq:dyn_expanded}
    \begin{aligned}
&\begin{bmatrix} \dot{X} & \dot{Y} & \dot{Z}\end{bmatrix}^\top = Q_{B/\mathbb{I}} \begin{bmatrix} v_x & v_y & v_z \end{bmatrix}^\top, \\
&\begin{bmatrix} \dot{\psi} & \dot{\theta} & \dot{\phi} \end{bmatrix}^\top = W^{-1} \begin{bmatrix} \omega_x & \omega_y & \omega_z \end{bmatrix}^\top, \\
&\begin{bmatrix} \dot{v}_x & \dot{v}_y & \dot{v}_z \end{bmatrix}^\top = \frac{1}{M}\begin{bmatrix} F_x & F_y & F_z \end{bmatrix}^\top - \Omega \begin{bmatrix} v_x & v_y & v_z \end{bmatrix}^\top,\\
& \begin{bmatrix} \dot{\omega}_x & \dot{\omega}_y & \dot{\omega}_z \end{bmatrix}^\top = J^{-1}\begin{bmatrix} \tau_x & \tau_y & \tau_z \end{bmatrix}^\top \\ & ~~~~~~~~~~~~~~~~~~~~~~~~~~~~~~~~~~ - J^{-1} \Omega J \begin{bmatrix} \omega_x & \omega_y & \omega_z \end{bmatrix}^\top,
    \end{aligned}
\end{equation}
with $M$ being the mass of the body and the angular velocity and rotation matrices given by
\begin{align*}
& \Omega = \begin{bmatrix} 0 & -\omega_z & \omega_y \\ \omega_z & 0 & -\omega_x \\ - \omega_y & \omega_x & 0 \end{bmatrix},~\textnormal{and}\\
    & Q_{B/I} = \begin{bmatrix} c \theta c \phi & c \phi s \theta s \phi - c \phi s \psi & s \phi s \psi + c \phi c \psi s \theta \\ c \theta s \psi & c \phi c \psi + s \theta s \phi s \psi & c \phi s \theta s \psi - c \psi s \phi \\ s \theta & c \theta s \phi & c \theta c \phi \end{bmatrix},
\end{align*}
respectively, where $\sin$ and $\cos$ have been abbreviated. Using \eqref{eq:dyn_expanded}, the state-space equation for the transported object is compactly written as: 
\begin{align}\label{eq:mod_con}
    \dot{S}(t) & = f_c(S(t), u(t)),
\end{align} 
with states and inputs at time $t$ given by: 
\begin{align*}
& S(t) = [X(t), Y(t), Z(t), \psi(t), \theta(t), \phi(t), v_x(t), v_y(t), v_z(t),\\ &~~~~~~~~~~~~~~~~~~~~~~~~~~~~~~~~~~~~~~~~~~~~~\omega_x(t), \omega_y(t), \omega_z(t)]^\top, \\ 
& u(t) = [F_x(t), F_y(t), F_z(t), \tau_x(t), \tau_y(t), \tau_z(t)]^\top.
\end{align*}
We discretize~\eqref{eq:mod_con} with the sampling time of $T_s$ of the robot to obtain its discrete time version: 
\begin{align}\label{eq:mod_gen}
    S_{t+T_s} = f(S_t, u_t).
\end{align} 
Given any input $u_t$ to the center of mass of the object, we decouple it into the corresponding human inputs $u^h_t$ and robot inputs $u^r_t$, such that $u_t = u^h_t + u^r_t$.   
% \subsection{Input Constraints}
We consider constraints on the inputs of the robot and the human given by $u^h_t \in \mathcal{U}^h$ and $u^r_t \in \mathcal{U}^r$ for all $t \geq 0$. The set $\mathcal{U}^h$ can be learned from human demonstrations' data. 
\section{Robot's Policy Design}
We detail the steps involved in control synthesis by the robot in this section. The robot computes the net (i.e., from both the human and the robot) optimal forces and torques to be applied to the center of mass of the transported body by solving a constrained finite time optimal control problem in a receding horizon fashion. The robot's portion of those net optimal inputs are affected by its proximity to obstacles potentially unknown to the human and an estimate of the human's assisting input. We elaborate these steps next. 

\subsection{MPC Planner and Human's Inputs Estimation}
The constrained finite time optimal control problem that the robot solves at time step $t$ with a horizon of $N \ll T$ is given by:
\begin{equation}\label{eq:generalized_InfOCP}
	\begin{aligned}
% V^{\star}(x_t,& \mathcal{P}_A, \mathcal{P}_B) = \notag \\
		\displaystyle\min_{U_t} & \displaystyle\sum\limits_{k = 1}^{N} [ (S_{t+kT_s|t} - S_{\mathrm{tar}})^\top Q_s (S_{t+kT_s|t} - S_{\mathrm{tar}}) + \cdots \\ & ~~~~~~~~~~~ + u_{t+(k-1)T_s|t}^\top Q_i u_{t+(k-1)T_s|t}] + \mathcal{I}_{\mathcal{O}}(S_t,u^h_{t-T_s}) 
		\\
		&\text{s.t.,}~~~~~{S}_{t+kT_s|t} = f(S_{t+(k-1)T_s|t}, u_{t+(k-1)T_s|t}), \\ % ~R_{k|t} = f_\mathcal{B}(S_{k|t}), \\ 
		&~~~~~~~~~ \mathcal{B}(S_{t+kT_s|t}) \in \mathcal{X} \setminus \mathcal{C}_{r,t},~u_{t+(k-1)T_s|t} \in \mathcal{U}^r \oplus \mathcal{U}^h,\\
% 		&~~~~~~~~~v_{t+(k-1)T_s|t} = \begin{bmatrix} \hat{f}^x_1(u_{t+(k-1)T_s|t}) \\ \hat{f}^y_1(u_{t+(k-1)T_s|t}) \\ \hat{\tau}(u_{t+(k-1)T_s|t})\end{bmatrix} \! \in \mathcal{V},\\
		&~~~~~~~~~\forall k \in \{1,2,\dots,N\},~S_{t|t} = {S}_t, % ~R_{t|t} = \hat{R}_t,
	\end{aligned}
\end{equation}
where $\mathcal{B}(\cdot)$ is a set of positions defining the transported object, 
% $f_\mathcal{B}(\cdot)$ is a function that relates the follower's center of mass position to the system states, 
$U_t = \{u_{t|t},\dots,u_{t+(N-1)T_s|t}\}$, $S_\mathrm{tar}$ is the target state, $Q_s, Q_i \succcurlyeq 0$ are the weight matrices, and inferred obstacle zone penalty $\mathcal{I}_{\mathcal{O}}(S_t, u^h_{t-T_s})$ is defined in Section~\ref{ssec:infer_obs}. Once an optimal input $u^\star_t$ is computed, the robot utilizes the following assumption to estimate the human's inputs. 
\begin{assumption}\label{assump:hum_input}
The human's compliant inputs at time step $t$ are computed as
\begin{align}\label{eq:human_input}
    \hat{u}^h_t = pu^\star_t,
\end{align}
where fraction $p \in (0,1)$ remains constant throughout the task. 
\end{assumption}
The fraction $p$ can be roughly estimated from collected trial data where the human limits to playing a complying role in the task\footnote{If the human actively leads the task, potentially forcing/opposing robot's actions, human inputs may be drastically different from its approximate \eqref{eq:human_input}.}. Thus, the robot's estimate of the human policy inherently considers that the human is trying to minimize their felt forces and torque in the task to assist the robot, while reacting to the surrounding obstacles in $\mathcal{C}_{r,t}$ in a way which is consistent with the MPC planned trajectory by the robot. Utilizing Assumption~\ref{assump:hum_input}, the robot computes its actions at $t$ as: 
\begin{align}\label{eq:rob_input}
    & u^{\star,r}_t = u^\star_t - \hat{u}^h_t.
\end{align}
%%%%%%%%%%%%%%%%%%%%%%
\subsection{Trust Value $\alpha_t$ via Difference in Estimated and Actual Human Behavior}
Since the robot does not perfectly know the human's intentions and the configuration of obstacles in the vicinity of the human, it does not apply its computed MPC input $u^{\star,r}_t$ to system \eqref{eq:mod_con} directly. Instead, it checks the deviation of its estimated human inputs from the actual closed-loop inputs applied by the human. The latter can be measured using force and torque sensors on the robot. As the applied human inputs at the current time step are not available for this computation, the robot approximates\footnote{For sample period $T_s \ll 1$, this can constitute a reasonable approximation.} this deviation by:
\begin{align*}
    \Delta u^h_t \approx \hat{u}^h_t - u^h_{t-T_s}. 
\end{align*}
The \emph{trust} value $\alpha_t$ is then computed as:
\begin{equation}\label{eq:alpha_update_equation}
    \begin{aligned}
    & \alpha_t  = 1-\min \{1, \frac{\Vert \Delta u^h_t\Vert} {\delta_\mathrm{thr}}\},
    \end{aligned}
\end{equation}
where $\delta_\mathrm{thr}$ is a chosen threshold deviation. The robot uses this trust value to apply a weighted combination of its computed MPC inputs $u^{\star, r}_{t}$, and inputs proportional to $u^h_{t-T_s}$ as detailed later in equation~\eqref{eq:mpc_pol_formulation}. This trust-driven combination of inputs is motivated by works such as \cite{dragan2013policy, nikolaidis2016formalizing,scobee2018haptic}. The robot additionally deploys a safe stopping policy, in case the computed trust value is below a chosen threshold, and it nears obstacles potentially unknown to the human. These two modes of the robot's policy are detailed in the next section. 
% We see two extreme cases as follows:
% \begin{itemize}
%     \item Big deviation then robot acts as a follower and lets the human take charge, until (C1) is triggered. Robot warns human and decelerates the object if it is nearing an obstacle and (C1) is active.
%     \item If no deviation then robot applies whatever it computes, as the actions reach a consensus.  
% \end{itemize}
%%%%%%%%%%%%%%%%%%%%%%%%%%%%%%%%%%%%%%%%%%%%%%%%%%%%%%
\subsection{Trust-Driven and Safe Stop Modes of the Robot Policy}
At time step $t$, we denote the inertial position coordinates of the robot's seen point on the object closest to any obstacle in $\mathcal{C}_{r,t}$ as $R_t$. After finding a solution to \eqref{eq:generalized_InfOCP} and computing $u^{\star,r}_t$ using \eqref{eq:rob_input}, the robot utilizes \eqref{eq:alpha_update_equation} and applies its closed-loop input computed as follows:
\begin{equation}
    \begin{aligned}\label{eq:mpc_pol_formulation}
    & u^r_t = \begin{cases} \mathrm{proj}_{\mathcal{U}^r}(\alpha_tu^{\star,r}_{t} + K_1(1-\alpha_t)  u^h_{t-T_s}),~\textnormal{if (SS) not true,} \\
    \mathrm{proj}_{\mathcal{U}^r}(-K_2 \frac{\dot{R}_{t}}{Ts}),~\textnormal{otherwise,}\\ \end{cases}
    \end{aligned}
\end{equation}
% \begin{equation}
%     \begin{aligned}\label{eq:v_thres_stuff}
%     & \big\vert \frac{ v_{t-T_s}}{v_{\mathrm{thres}}} \big\vert v_{t-T_s}
%     \end{aligned}
% \end{equation}
to system \eqref{eq:mod_gen} in closed-loop with chosen gains $K_1, K_2 >0$, where $\mathrm{proj}_\mathcal{A}(x)$ denotes the Euclidean projection of $x$ onto set $\mathcal{A}$, and the robot's safe stop policy triggering condition (SS) is given by:
\begin{align}\label{eq:SS}
   \textnormal{(SS)}: \alpha_t \! < \frac{1}{2},~\min_{o \in \mathcal{C}_{r,t}} \Vert R_{t} - o \Vert \leq d_\mathrm{thr},\dot{R}_{t} \cdot (o - R_{t})> v_\mathrm{thr},
\end{align}
with distance and velocity thresholds $d_\mathrm{thr}>0$ and $v_\mathrm{thr}>0$. That is, when point $R_t$ approaches any obstacle $o$ at a high velocity under a low trust $\alpha_t < \frac{1}{2}$, the robot actively tries to decelerate the the object and bring it to a halt. From policy \eqref{eq:mpc_pol_formulation}, we make the following observations: 
% We see two extreme cases as follows:
\begin{enumerate}
\item  A large trust value (e.g., $\alpha_t$ closer to 1), corresponds to the case when the robot's estimates of the human's inputs align with the actual human's inputs. This means that the human is taking on a  follower role, trusting on the robot's actions. The robot trusts the computed inputs $u^{\star,r}_{t}$ from the MPC problem \eqref{eq:generalized_InfOCP} and takes the leader's role in the task. 

\item A small trust value (e.g., $\alpha_t$ close to 0) corresponds to the case when the robot's predictions of the human's inputs do not align with the actual human's inputs. This means that the human is taking on the leader's role, either reacting to obstacles nearby or actively leading the task. The robot does not trust the computed inputs $u^{\star, r}_{t}$ from the MPC problem \eqref{eq:generalized_InfOCP} and takes the follower's role (unless the safe stop policy condition is triggered).
\end{enumerate}
Policy~\eqref{eq:mpc_pol_formulation} is motivated by \cite{sadrfaridpour2018trust}, and qualitatively has the properties of joint impedance and admittance, similar to \cite{ott2015hybrid}. We see that satisfying condition 1 increases the efficacy of the robot's solution to \eqref{eq:generalized_InfOCP}, i.e., $u^{\star, r}_{t}$. To that end, we add the inferred obstacle zone penalty $\mathcal{I}_\mathcal{O}(S_t, u^h_{t-T_s})$ to~\eqref{eq:generalized_InfOCP}, adapting the cost to be optimized by inferring information on potential obstacles at the human's vicinity. This is elaborated next. 
\subsection{Increasing Trust $\alpha_t$ via Inferred Obstacle Zone Penalty $\mathcal{I}_{\mathcal{O}}(S_t, u^h_{t-T_s})$}\label{ssec:infer_obs}
At time step $t$, we denote the inertial position coordinates of the human by $H_t$. We also denote the first three force components of the human input $u^h_t$ by $u^h_{f,t}$. Motivated by the obstacle learning work of \cite{bujarbaruah2021decentralized}, we add the extra term $\mathcal{I}_{\mathcal{O}}(S_t, u^h_{t-T_s})$ to the cost in \eqref{eq:generalized_InfOCP} at every time step. This term is to be chosen when $\alpha_t < \frac{1}{2}$, and the human applies forces along directions which are more than a user specified threshold $\nu_\mathrm{thr}$ radians apart from its expected ones. We then choose the term $\mathcal{I}_{\mathcal{O}}(S_t, u^h_{t-T_s})$ as follows:
\begin{equation}
    \begin{aligned}\label{eq:io}
    &\mathcal{I}_{\mathcal{O}}(S_t, u^h_{t-T_s}) = \begin{cases} \sum_{i=1}^{n} \frac{1}{\Vert (S_t - H_t + K_3 u^h_{f,t-T_s} + o_i)\Vert},~\textnormal{if (IO)}, \\ 0,~\textnormal{otherwise}, \end{cases}
    \end{aligned}
\end{equation}
with $n$ choices of the random parameter $0<o_i\ll 1$ (introduces noise in the direction vector), control gain $K_3>0$, and condition (IO) being 
\begin{align*}
    \textnormal{(IO)}: \alpha_t < \frac{1}{2},~ \vert \arccos (\frac{\hat{u}^h_{f,t} \cdot u^h_{f,t-T_s}}{\Vert \hat{u}^h_{f,t}\Vert \Vert u^h_{f,t-T_s}\Vert}) \vert > \nu_\mathrm{thr}. 
\end{align*}
% \begin{remark}
Intuitively, we assume that if the human unexpectedly pushes against the robot, they are attempting to avoid some obstacle unknown to the robot. The robot uses these force measurements and generates $n$ virtual obstacle points that are placed relative to the human's location at a distance scaled by the negative force vector, plus some noise. These virtual obstacle points are the robot's estimates of potential obstacles in the human's vicinity, due to which the human's input $u^h_{t-T_s}$ is significantly different from the estimate $\hat{u}^h_t$. Introducing the penalty $\mathcal{I}_\mathcal{O}(S_t, u^h_{t-T_s})$ can improve the MPC planner \eqref{eq:generalized_InfOCP}, increasing the value of $\alpha_t$ and enabling more effective role of the robot in the task.     
% \end{remark}
%%
\section{Experimental Results}\label{sec:numerics}
In this section, we present experimental validation results with our proposed approach. The experiments are conducted with a UR5e robot. 
% The human and the robot start the joint transportation task with the center of mass of the transported rigid box at the start state $S_0$, as shown in Fig.~\ref{fig:exp_space}. Goal state $S_\mathrm{tar}$ contains the target location, which is known to both agents. 
Since there is not an exact shared baseline for this problem formulation of a human-robot collaborative transportation task with partial obstacle information, we avoid directly comparing against controllers from other related work. We use the following set of parameters shown in Table~\ref{table:params} for the considered experimental scenario.
\begin{table}[h!]
\centering
\caption{Parameters used in control design.}
\label{table:params}
\begin{tabular}{|c|c|c|} 
 \hline
 Parameter & Value\\
 \hline
 $T, T_s$ & 100s, 0.05s\\
 $N$ & 20\\
 $p$ & 0.5 \\ 
 $d_\mathrm{thr}, v_\mathrm{thr}, \nu_\mathrm{thr}$ & 0.15m, 0.05m/s, $\frac{\pi}{6}$rad \\
%  $\nu_\mathrm{thr}$ & $\frac{\pi}{6}$rad\\
 $K_1, K_2, K_3$ & 1, 10, 0.005\\
 $Q_s$ & $\mathrm{diag}$(20,20,20,1,1,1)\\
 $Q_i$ & $\mathrm{diag}$(10,10,10,100,100,100)\\
 \hline
\end{tabular}
\end{table}
\subsection{Trust-Driven Policy vs Pure MPC Policy}\label{sec:IO}
\begin{figure*}[h]
% \captionsetup[subfigure]{}
\centering

    \subfloat[Pure MPC policy ($\alpha = 1$) resulting in collision with obstacle only known by the human.]{%
        \includegraphics[width=0.9\columnwidth]{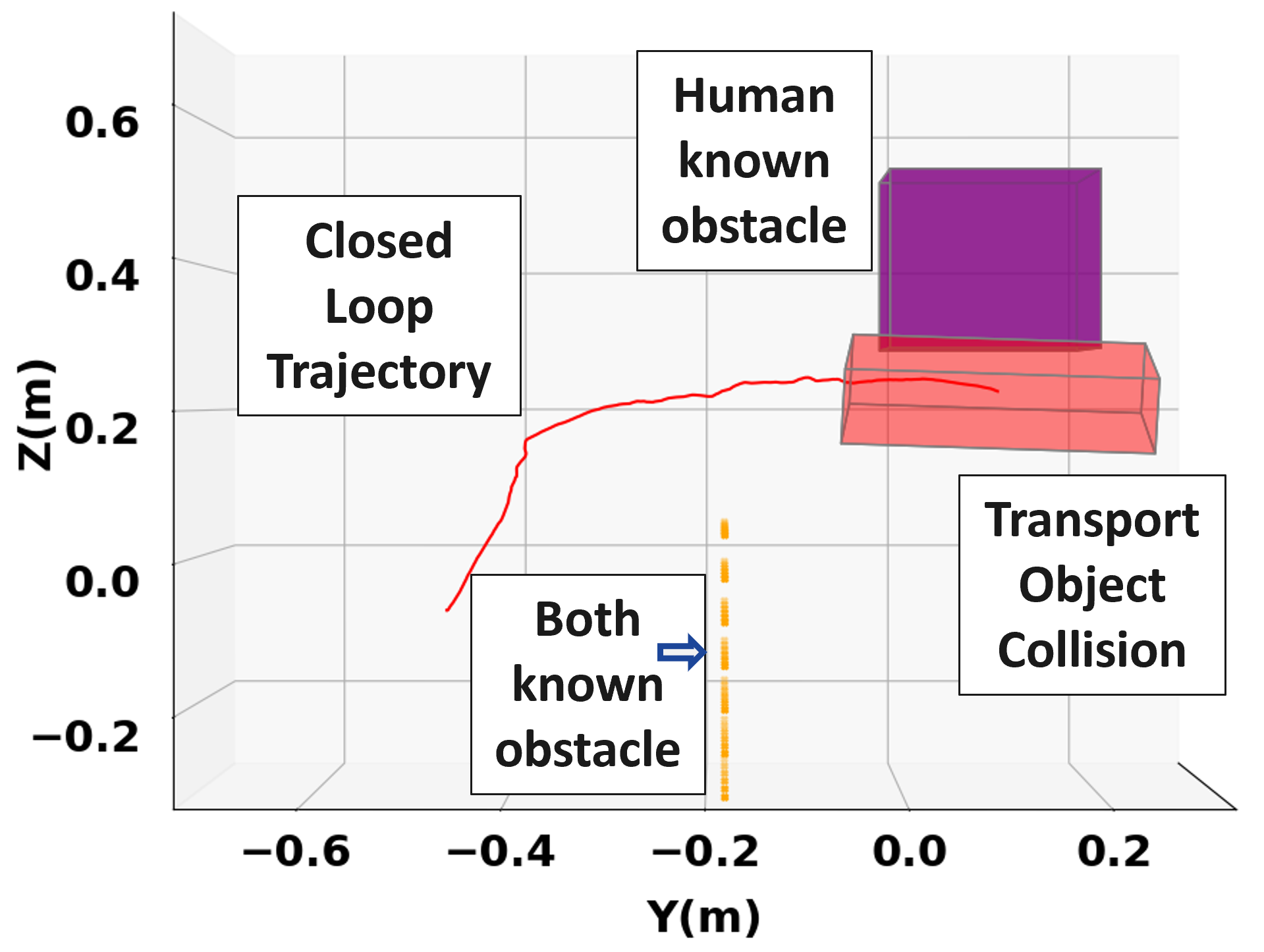}
        \label{fig:no_alpha}
    }
    \hspace{1em}
    \subfloat[Trust-Driven Policy (adaptive $\alpha$) resulting in a collision-free trajectory.]{%
        \includegraphics[width=0.9\columnwidth]{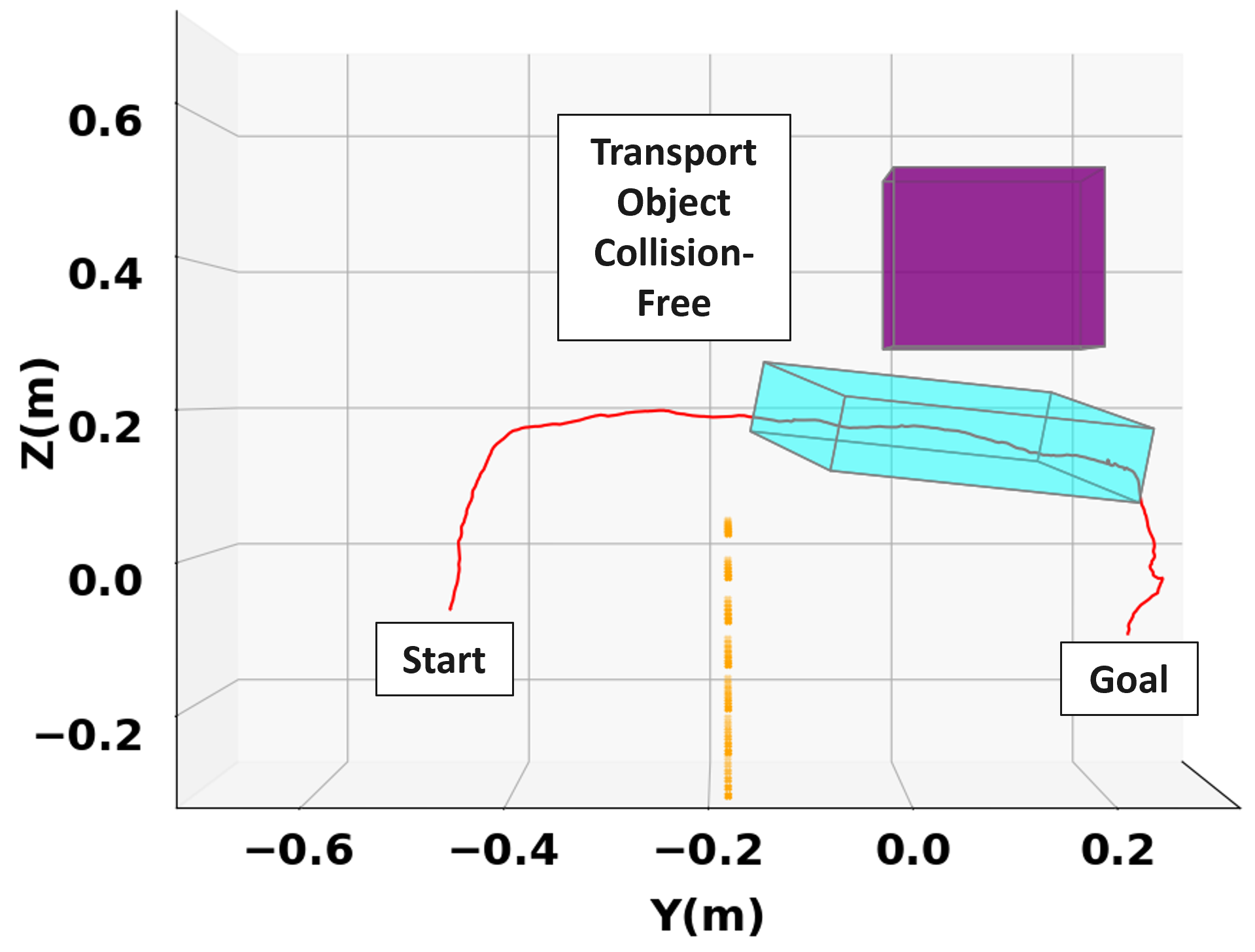}
        \label{fig:alpha}
    }
    \\
    \subfloat[$\alpha$ vs Time. The trust-driven policy adapts the value of $\alpha_t$ for all $t \geq 0$ based on the human's responses in the task.]{%
        \includegraphics[width=0.9\columnwidth]{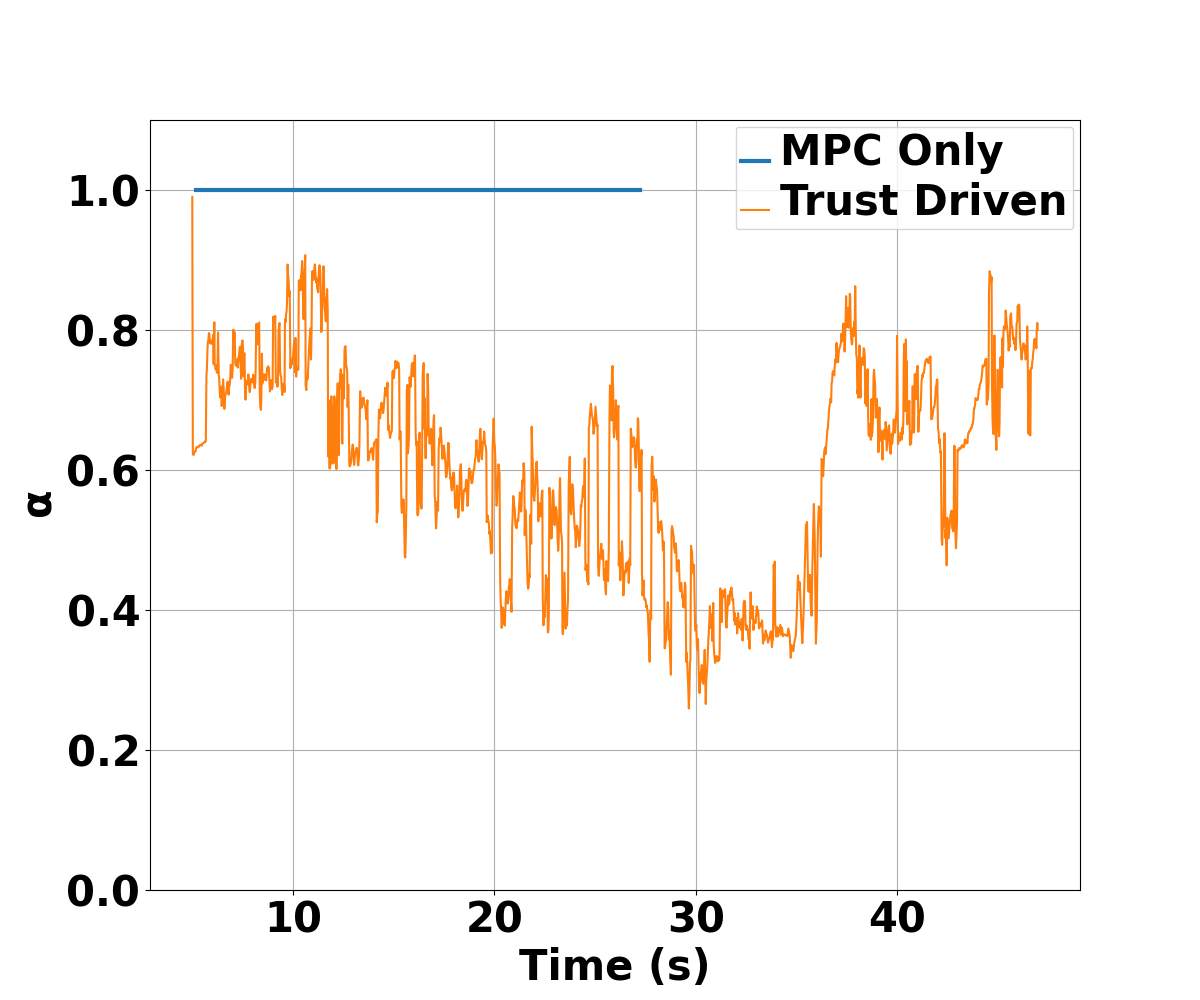}
        \label{fig:alpha_n_alpha_alpha}
    }
    \hspace{1em}
    \subfloat[Measurement of human force applied in Z direction vs Time. Using the trust-driven policy enables the human to lower resisting forces, while avoiding collision.]{%
        \includegraphics[width=0.85\columnwidth]{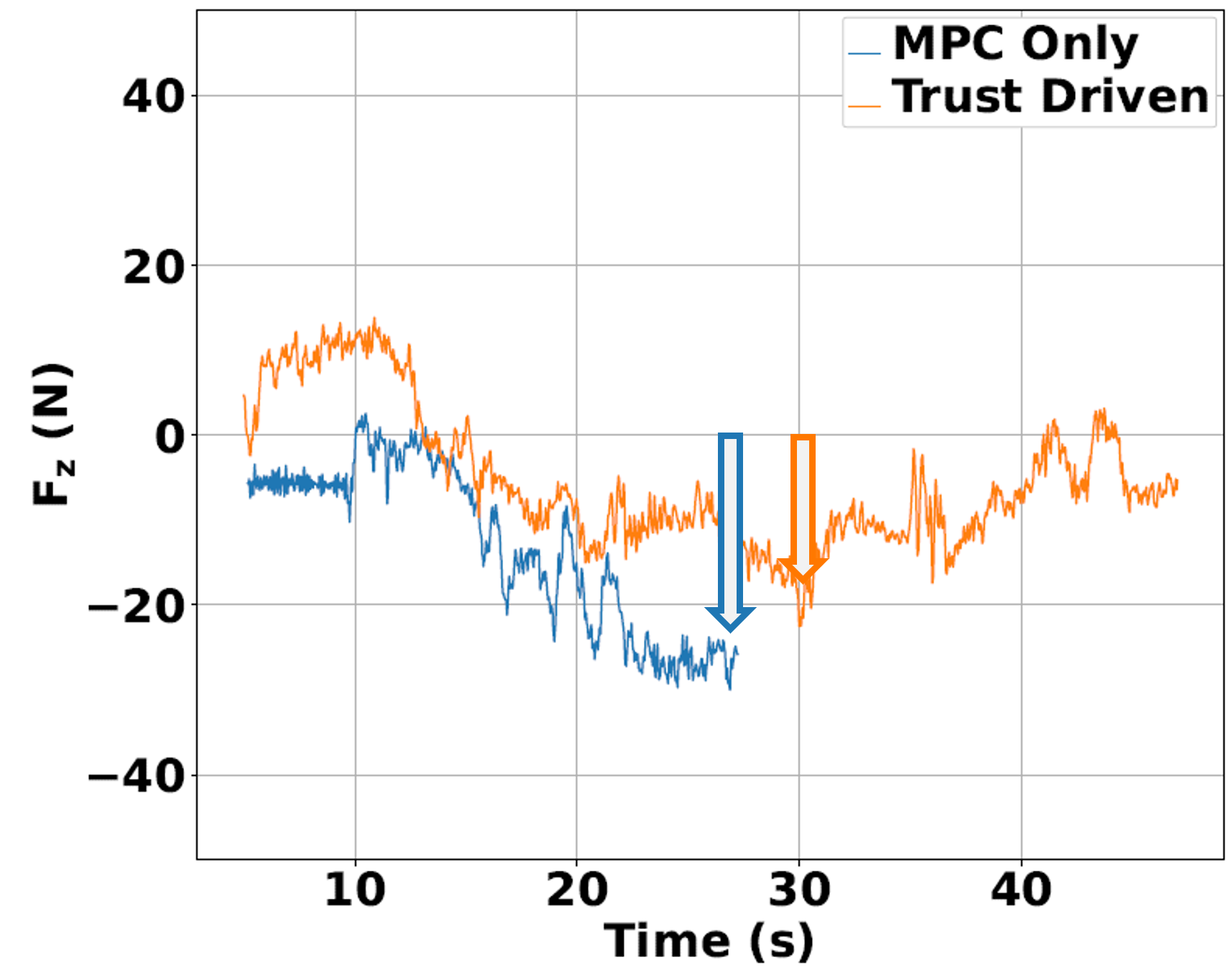}
        \label{fig:alpha_n_alpha_forces}
    }
    \caption{Comparison of experimental results of robot with pure MPC policy vs. trust-driven policy.}
    \label{fig:trust_pol_ben}
\end{figure*}

For this section, two obstacles are placed between the agents and the target, as shown in the rendered experiment space in Fig.~\ref{fig:alpha}. We first show the benefits of using the trust-driven policy mode, where the robot utilizes the trust value $\alpha_t$ to apply a weighted combination of its MPC inputs and the human's inputs to the system. The baseline for comparison is a pure MPC policy, with the robot solving MPC problem \eqref{eq:generalized_InfOCP} and applying its optimal input \eqref{eq:rob_input}, being agnostic to the responses of the human. 
In the considered scenario in Fig.~\ref{fig:trust_pol_ben}, the purple box obstacle located between $Z \in [0.35m, 0.57m]$ is known only by the human. Both agents are aware of the dotted wall obstacle at $Y = -0.2$ m. In Fig.~\ref{fig:no_alpha}, the robot is operating with the pure MPC baseline policy, agnostic to the human's actions. As a consequence, the planned trajectory by the robot results in the human colliding with this obstacle, as seen in Fig.~\ref{fig:no_alpha}. Resisting force values by the human in Fig.~\ref{fig:alpha_n_alpha_forces} indicate the human's opposition to the robot's actions. On the other hand, with our proposed trust-driven policy mode, the robot is cognizant of the human's intentions. The evolution of $\alpha_t$ as the human navigates in the proximity of the box obstacle is shown in Fig.~\ref{fig:alpha_n_alpha_alpha}. When the transport object nears the obstacle (around 30 sec), the robot distrusts its estimate of the human policy with a computed $\alpha_t \approx 0.3$ and applies more of the measured human input in ~\eqref{eq:mpc_pol_formulation}. Collision is averted as a consequence, as seen in Fig.~\ref{fig:alpha}. Lower force magnitudes in Fig.~\ref{fig:alpha_n_alpha_forces} further indicate that the human's resistance to robot's actions during this collision avoidance is lowered, as the robot lowers the contribution of its MPC inputs in \eqref{eq:mpc_pol_formulation} with a low value of $\alpha_t$. 

\subsection{The Safe Stop Mode in Action}
\begin{figure*}[h]
% \captionsetup[subfigure]{}
\centering

    \subfloat[No safe stop policy. The robot collides with the obstacle.]{%
        \includegraphics[width=0.9\columnwidth]{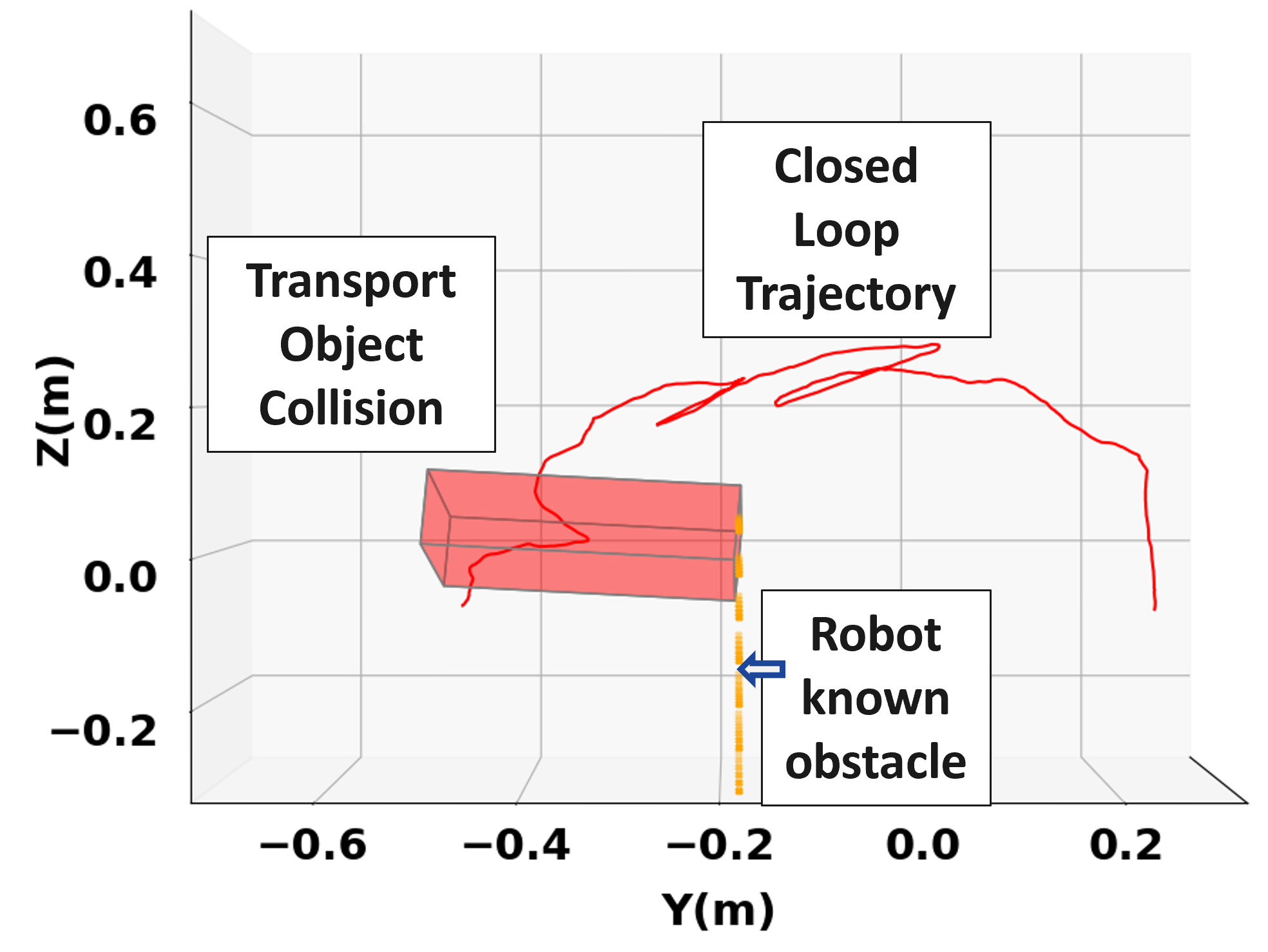}
        \label{fig:ss1}
    }
    \hspace{1em}
    \subfloat[With safe stop policy. Collision is avoided. ]{%
        \includegraphics[width=0.9\columnwidth]{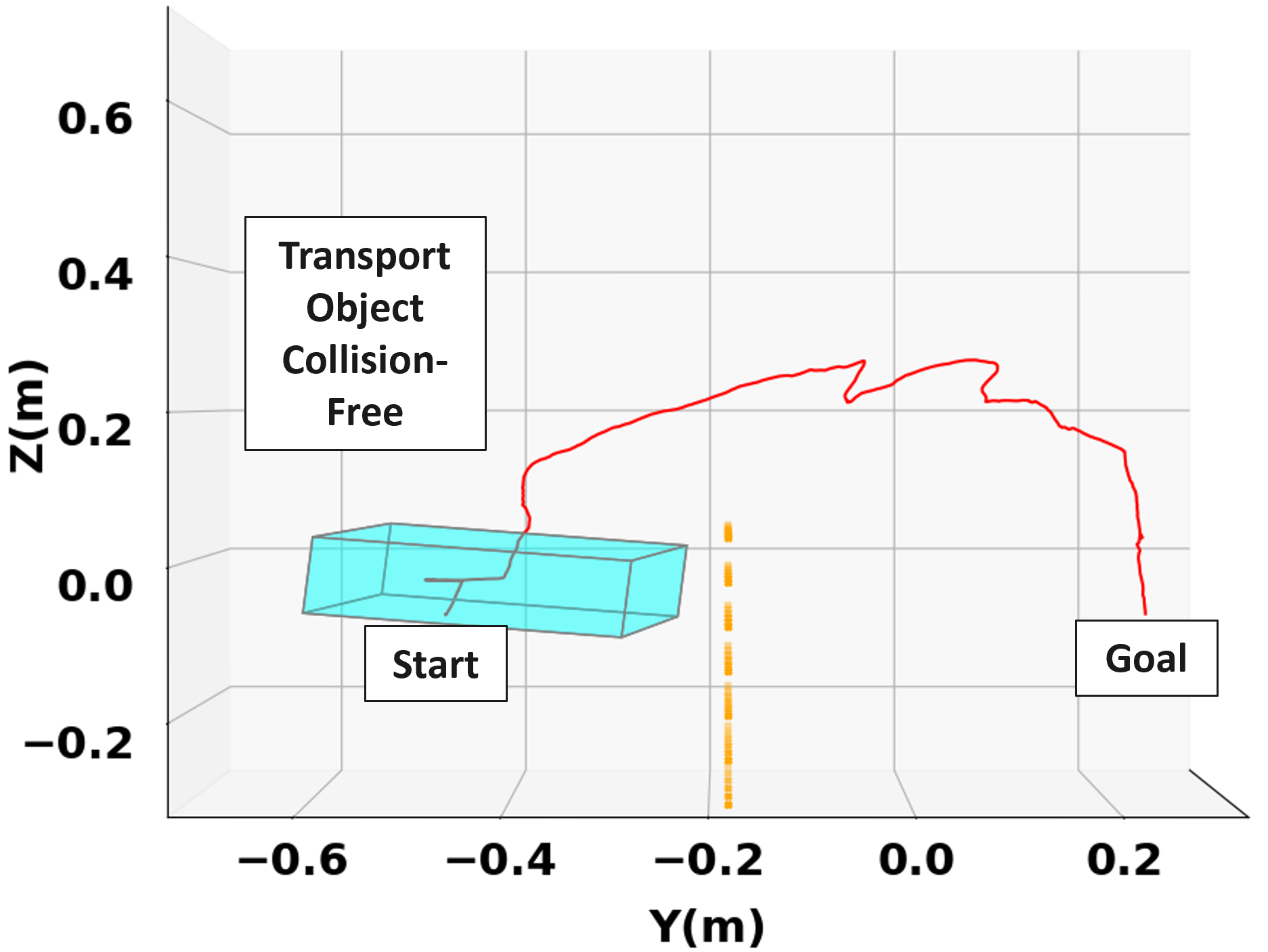}
        \label{fig:ss2}
    }
    \\
    \subfloat[Without the safe stop policy, the robot provides assisting force that matches the unexpected human inputs even if it leads towards a known obstacle (marked from 8.8s to 11.9s). The human behavior causes a collision with the obstacle wall and the robot helps them do so. ]{%
        \includegraphics[width=0.9\columnwidth]{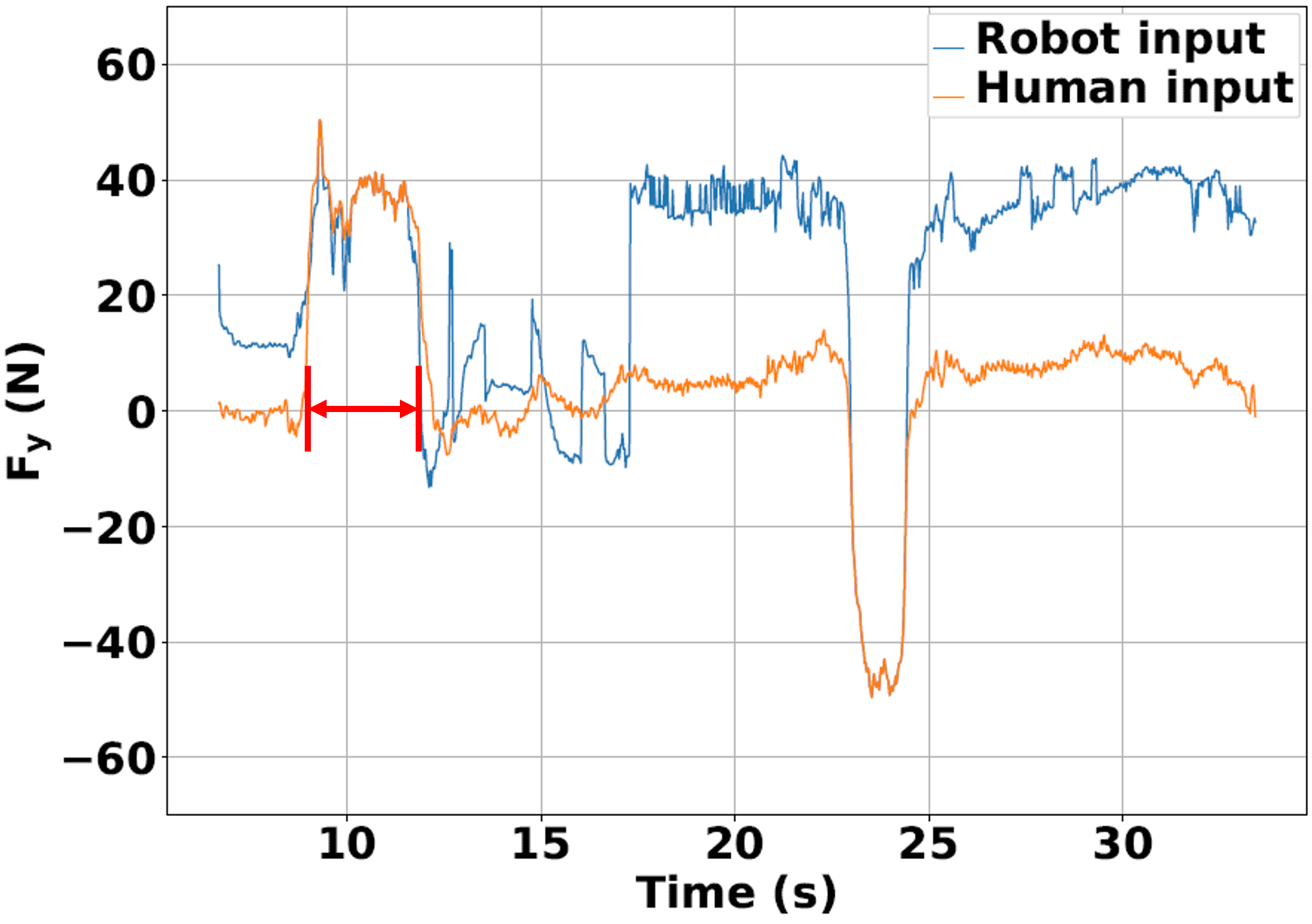}
        \label{fig:ss3}
    }
    \hspace{1em}
    \subfloat[With the safe stop policy, the robot applies decelerating safe stop input to cancel out the human inputs when it detects that a collision with an obstacle is imminent (marked from 8.5s to 11.2s). This prevents the human from leading the transport object into the obstacle wall]{%
        \includegraphics[width=0.95\columnwidth]{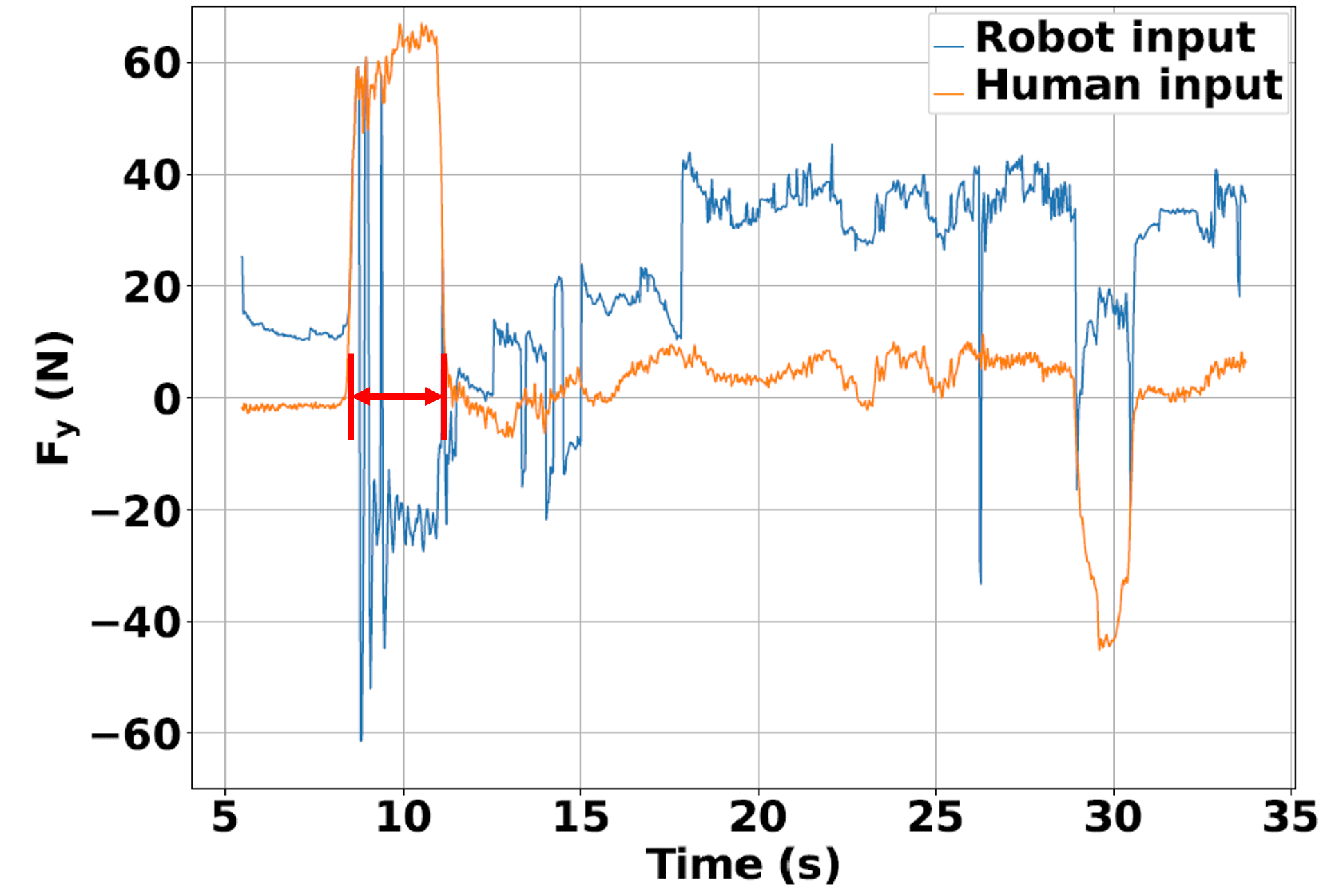}
        \label{fig:ss4}
    }
    \caption{Effect of the safe stop policy mode in avoiding collisions.}
    \label{fig:ss}
\end{figure*}

To highlight the safety benefits of adding the safe stop policy mode in \eqref{eq:mpc_pol_formulation}, we consider the scenario shown in Fig.~\ref{fig:ss}. For this scenario, only one simulated obstacle wall at $Y = -0.2$ m is in the experiment space which the human does not see. The human decides to drive the transport object towards the goal via the shortest path without being aware that it is leading towards the wall.
Without activating the safe stop policy backup, the robot's inputs continue to comply with the inputs from the human, as shown in the force plots in Fig.~\ref{fig:ss3}. As a result, the transported object collides with the obstacle wall, as seen in Fig.~\ref{fig:ss1}. On the other hand, in Fig.~\ref{fig:ss2} we see that utilizing the safe stop policy mode manages to prevent this collision and maintain safety in the transportation task. This safety retaining effect of the safe stop mode can be explained from Fig.~\ref{fig:ss4}, where next to the obstacle wall when condition (SS) is triggered (around 10 sec), we no longer see the robot's applied forces complying with the human's forces. Instead, the robot applies a decelerating safe stop input, which results in the collision avoidance. The task is completed successfully.   

\subsection{Randomized Analysis}
In order to generalize the validity of the above results beyond the considered example, we carried out the transportation task and analyzed the closed loop behaviors of the proposed controller with 100 configurations of randomized start, goal and obstacle positions. In some cases, the obstacles are purely simulated for faster testing purposes. The detailed results are shown in Table~\ref{table:results} where we use three metrics to compare the 100 trials. A \emph{Collision-Free Success} is a trial where the transport object is brought to the target state without hitting obstacles. \emph{Peak Human Force} is the largest magnitude of force applied by the human throughout a given trial. The \emph{Duration of Intervening Forces} is the length of time in which the human has applied more than 30N in a given trial.
% , where we use the definitions: 
% \begin{align*}
%     &u^h_M = \max_{t=0}^T \Vert u_t \Vert_2,~t_f = \sum_{t=0}^T tI(\Vert u_t \Vert_2 > 30\textnormal{N}),
% \end{align*}
% where the indicator function:
% \begin{align*}
%     I(S) = \begin{cases} 1,~\textnormal{if statement $S$ is true}, \\ 0,~\textnormal{otherwise.}\end{cases}
% \end{align*}
\begin{table}[h!]
\centering
\caption{The percentage and the average are computed numerically from 100 trials of the transportation task. }
\label{table:results}
\begin{tabular}{ |c|c|c|c|} 
 \hline
 Feature & MPC Only & \begin{tabular}{@{}c@{}} Trust-Driven \\ w/ Safe Stop\end{tabular}  \\
 \hline
 Collision-Free Successes ($\%$) & 51 & 88\\ 
 Avg. Peak Human Force  (N) & 63.276 & 53.835 \\
%  Std. Peak Human Force (N) & 28.767 & 35.043 \\
 Avg. Duration of Intervening Forces (s) & 5.934 & 2.265 \\
%  Std. Duration of Intervening Forces (s) & 5.242 & 2.404 \\
%  Avg. Time in a CFT (s) & 51.877 & 66.602 \\
 \hline
\end{tabular}
% \vspace{-18pt}
\end{table}
Table~\ref{table:results} shows that the proposed approach results in a 37\% increase in the number of Collision-Free Successes. Moreover, the average value of the Peak Human Force lowers by 14.9\% with the proposed approach, indicating decreased opposition of the human during the task. The results show that the average Duration of Intervening Forces shortens by 61.8\% with our approach. The robot cedes some of the control authority to the human as the trust value decreases. This occurs when the human does something unexpected to the robot. On the other hand, with the pure MPC approach, the robot attempts to follow its optimal trajectory even in the case where a collision with an object known only by the human is imminent. Thus, the human needs to continuously apply the intervening force for longer periods of time when no trust value is used.
%%%
\section{Conclusion}
We proposed a framework for a human-robot collaborative transportation task in presence of obstacles in the environment. The robot plans a trajectory for the transported object by solving a constrained finite time optimal control problem and appropriately applies a weighted combination of the human's applied and its own planned inputs. The weights are chosen based on the robot's trust value on its estimates of the human's inputs. This allows for a dynamic leader-follower role adaptation of the robot throughout the task. With experimental results, we demonstrated the efficacy of the proposed approach. 

% Acknowledgments---Will not appear in anonymized version
\section*{Acknowledgments}
We thank Vijay Govindarajan and Conrad Holda for all the helpful discussions. This work was funded by ONR-N00014-18-1-2833, and NSF-1931853. This work is also supported by AFRI Competitive Grant no. 2020-67021-32855/project accession no. 1024262 from the USDA National Institute of Food and Agriculture. This grant is being administered through AIFS: the AI Institute for Next Generation Food Systems (\href{https://aifs.ucdavis.edu}{https://aifs.ucdavis.edu}).
\nocite{*}
% \newpage% Generated by IEEEtran.bst, version: 1.14 (2015/08/26)

\end{document}